\documentclass{article}

\usepackage{microtype}
\usepackage{graphicx}
\usepackage{subcaption}
\usepackage{booktabs}
\usepackage{multirow}
\usepackage{amsmath}
\usepackage{amssymb}
\usepackage{mathtools}
\usepackage{algorithm}
\usepackage{algorithmic}
\usepackage{hyperref}

\usepackage[accepted]{icml2026}
\usepackage[capitalize,noabbrev]{cleveref}

\icmltitlerunning{Boosting DPO with Penalization}

\begin{document}

\twocolumn[
  \icmltitle{Boosting Direct Preference Optimization with Penalization}

  \begin{icmlauthorlist}
    \icmlauthor{Pengwei Sun}{anon}
  \end{icmlauthorlist}
  \icmlaffiliation{anon}{Stanford University}
  \icmlcorrespondingauthor{Pengwei Sun}{pengwei@stanford.edu}
  \icmlkeywords{RLHF, preference optimization, DPO, instruction tuning}

  \vskip 0.3in
]

\printAffiliationsAndNotice{}

\begin{abstract}
Offline preference optimization has become a practical substitute for reinforcement learning from human feedback, but pairwise objectives such as Direct Preference Optimization (DPO) and its variants use only the chosen and rejected responses stored in a static dataset. This leaves a useful signal unused: the response that the reference model itself would generate for the same prompt. We propose Direct Preference Optimization with Penalization (DPOP), a simple extension of DPO that augments the base preference loss with a gated penalty on reference-greedy responses. DPOP activates this penalty only when the current policy still assigns a lower likelihood to the preferred response than to the rejected response. On AlpacaEval 2.0, DPOP improves length-controlled win rate over DPO, SimPO, and AlphaDPO on both Llama-3-8b-it and Gemma-2-9b-it, achieving relative gains of 5.3\% and 4.4\% over baselines on the two models, respectively. Ablations further show that a SimNPO-style length-normalized penalty is stronger than NPO and token-level unlikelihood in this setting.

\end{abstract}

\section{Introduction}

Aligning large language models with human preferences has traditionally relied on Reinforcement Learning from Human Feedback (RLHF), a complex, multi-stage pipeline requiring a separately trained reward model and constrained policy optimization \citep{ouyang2022training}. Direct Preference Optimization (DPO) simplified this pipeline by bypassing the explicit reward model entirely, elegantly re-framing alignment as a purely pairwise classification task over reference-corrected policy likelihoods \citep{rafailov2023direct}. This formulation has made offline alignment substantially easier to implement, and it remains a strong baseline because it preserves the regularizing role of a reference model without requiring online rollouts during training.

The same simplification also imposes a restriction: the training signal is confined to the two responses already stored in the preference dataset. This is limiting for modern instruction tuning, where preference data are often off-policy and are frequently produced or filtered by teacher models rather than collected directly from the policy being trained. In this regime, the reference model is not merely a source of KL regularization. It is also a concrete generator whose own greedy response to each prompt can reveal local modes that the policy may continue to imitate unless explicitly corrected. But DPO only updates the policy on $(x,y_w,y_l)$ and leave the reference-greedy output $y_g$ outside the objective.

We introduce Direct Preference Optimization with Penalization (DPOP), a minimal offline preference objective that incorporates this missing response. DPOP keeps the ordinary pairwise preference loss as the base objective, but additionally penalizes the policy likelihood of the reference-greedy response under a detached policy-margin gate. The gate is active only when the policy still ranks the rejected response above the chosen response in likelihood, which targets examples where the current policy has not yet internalized the pairwise preference.

Empirically, this design is especially effective with a SimNPO-style penalty, which is length-normalized and reference-free on the penalized response. On AlpacaEval 2.0, DPOP improves the length-controlled win rate (LC-WR) from 44.01 for SimPO and 41.48 for AlphaDPO to 46.35 on Llama-3-8b-it. On Gemma-2-9b-it, it improves from 73.08 for SimPO and 74.90 for AlphaDPO to 78.22. These results suggest that reference-greedy responses are not only useful as cached artifacts for analysis, but can be turned into an effective offline penalization signal.

\section{Related Work}

\textbf{Direct preference optimization.}
DPO uses a binary preference loss derived from the Bradley--Terry model \citep{bradley1952rank,rafailov2023direct}, optimizing $\mathcal{L}_{\mathrm{DPO}} = -\mathbb{E}_{(x, y_w, y_l)} \left[ \log \sigma\left(\beta \left[\log \frac{\pi_\theta(y_w|x)}{\pi_{\mathrm{ref}}(y_w|x)} - \log \frac{\pi_\theta(y_l|x)}{\pi_{\mathrm{ref}}(y_l|x)}\right]\right) \right]$. Given a prompt, chosen response, and rejected response, DPO compares the policy's log-likelihood ratio with the corresponding reference-model log-likelihood ratio. Its stability and simplicity make it the most natural base loss for DPOP. Our method preserves this pairwise DPO component, but adds a separate penalty on the response generated greedily by the reference model.

\textbf{Reference-free and adaptive-margin objectives.}
Several variants of DPO have been developed to improve its margin and reference mechanisms. SimPO argues that sequence-average log probability is better aligned with inference than a reference-corrected implicit reward, and removes the reference model from the preference loss while adding a fixed target margin $\gamma$ \citep{meng2024simpo}, optimizing \[
\begin{aligned}
\mathcal{L}
&=
-\mathbb{E}_{(x, y_w, y_l)}
\Bigg[
\log \sigma \Bigg(
\beta \Bigg[
\frac{1}{|y_w|}\log \pi_\theta(y_w \mid x) \\
&\qquad\qquad\qquad
-
\frac{1}{|y_l|}\log \pi_\theta(y_l \mid x)
-
\gamma
\Bigg]
\Bigg)
\Bigg].
\end{aligned}
\] AlphaDPO instead keeps a reference-based perspective but adapts the reward margin according to an interpolation between policy and reference behavior \citep{wu2024alphadpo}, optimizing\[
\begin{aligned}
\mathcal{L}
&=
-\mathbb{E}_{(x, y_w, y_l)}
\Bigg[
\log \sigma \Bigg(
\frac{\beta}{|y_w|}
\log \pi_\theta(y_w \mid x) \\
&\qquad\qquad\qquad
-
\frac{\beta}{|y_l|}
\log \pi_\theta(y_l \mid x)
-
sg(\gamma + \alpha M)
\Bigg)
\Bigg],
\end{aligned}
\]
where \[
M(x, y_w, y_l)
=
\beta
\left(
\log
\frac{
\pi_\theta(y_w \mid x)\pi_{\mathrm{ref}}(y_l \mid x)
}{
\pi_{\mathrm{ref}}(y_w \mid x)\pi_\theta(y_l \mid x)
}
\right).
\]
\textbf{Negative preference optimization.}
NPO was introduced for machine unlearning as a preference-style alternative to direct gradient ascent on undesirable data $y$ \citep{zhang2024npo}, optimizing $\mathcal{L}_{\mathrm{NPO}} = -\mathbb{E}_{(x, y) \sim \mathcal{D}} \left[ \frac{2}{\beta}\log \sigma\left(-\beta \log \frac{\pi_\theta(y|x)}{\pi_{\mathrm{ref}}(y|x)}\right) \right]$. SimNPO later revisited this setting and showed that removing the reference model from the negative objective can improve stability and effectiveness \citep{fan2024simnpo}, optimizing $\mathcal{L}_{\mathrm{SimNPO}} = -\mathbb{E}_{(x, y) \sim \mathcal{D}} \left[ \frac{2}{\beta}\log \sigma\left(-\beta \frac{1}{|y|}\log \pi_\theta(y|x) - \gamma\right) \right]$. DPOP adapts these penalty families to preference alignment rather than unlearning. The reference-greedy response is not private or harmful data; it is a model-generated response that may represent an undesirable local mode for the current prompt. This makes NPO and SimNPO useful penalty templates inside an otherwise standard preference-optimization pipeline.

\section{Methodology}

\subsection{Base Preference Objective}

Let $x$ denote a prompt, $y_w$ a preferred response, and $y_l$ a rejected response. Let $\pi_\theta$ be the trainable policy and $\pi_{\mathrm{ref}}$ the fixed reference model. The DPO base loss uses the reference-corrected logit
\begin{equation}
z_{\mathrm{dpo}}
=
\log
\frac{
\pi_\theta(y_w \mid x)\pi_{\mathrm{ref}}(y_l \mid x)
}{
\pi_{\mathrm{ref}}(y_w \mid x)\pi_\theta(y_l \mid x)
}.
\end{equation}
and optimizes
\begin{equation}
\mathcal{L}_{\mathrm{base}}(x,y_w,y_l)
= -\log \sigma(\beta z_{\mathrm{dpo}}),
\end{equation}
where $\beta$ is the preference temperature.

\subsection{Reference-Greedy Penalization}

DPOP augments each preference triple with a reference-greedy response
\begin{equation}
y_g = \mathrm{GreedyDecode}\big(\pi_{\mathrm{ref}}(\cdot|x)\big).
\end{equation}
This response is generated offline and cached in the training data. The core design choice is to penalize $y_g$ selectively, not universally. We first compute the policy likelihood margin
\begin{equation}
r = \mathrm{sg}\left(\log \pi_\theta(y_w|x)-\log \pi_\theta(y_l|x)\right),
\end{equation}
where $\mathrm{sg}(\cdot)$ denotes stop-gradient. The penalty gate is
\begin{equation}
g_{\pi}=\mathrm{sg}\left(\mathbf{1}[r<0]\right),
\end{equation}
so the reference-greedy penalty is active only when the policy still assigns higher likelihood to the rejected response than to the chosen response. DPOP then defines a detached penalty weight
\begin{equation}
w_{\mathrm{pen}} = g_{\pi} f(r),
\end{equation}
where $f$ is chosen from simple nonnegative families: linear, $f(r)=\max(-r,0)$; constant; square root, $f(r)=\sqrt{\max(-r,0)+\epsilon}$; and quadratic, $f(r)=\max(-r,0)^2$.

\subsection{Penalty Families}

We consider three penalties on $y_g$. Token-level unlikelihood penalizes the next-token probabilities assigned by the policy to the reference-greedy response:
\begin{equation}
\mathcal{L}_{\mathrm{unll}}
=-\frac{1}{|y_g|}\sum_t \log\left(1-\pi_\theta(y_{g,t}|x,y_{g,<t})\right).
\end{equation}
The NPO penalty uses a reference-relative negative objective,
\begin{equation}
\mathcal{L}_{\mathrm{npo}}
=-\frac{2}{\beta_p}\log \sigma\left(-\beta_p
\left[\log \pi_\theta(y_g|x)-\log \pi_{\mathrm{ref}}(y_g|x)\right]\right),
\end{equation}
where $\beta_p$ is the penalty temperature. The SimNPO penalty removes the reference from the negative objective and uses length-normalized log probability:
\begin{equation}
\mathcal{L}_{\mathrm{simnpo}}
=-\frac{2}{\beta_p}\log \sigma\left(-\beta_p \overline{\log \pi_\theta(y_g|x)}-\gamma_p\right),
\end{equation}
where $\overline{\log \pi_\theta}$ denotes average sequence log probability and $\gamma_p$ is a penalty margin.

The final DPOP objective is
\begin{equation}
\mathcal{L}_{\mathrm{DPOP}}
=\mathcal{L}_{\mathrm{base}} + 
\lambda_{\mathrm{pen}} w_{\mathrm{pen}}\mathcal{L}_{\mathrm{pen}},
\end{equation}
where $\mathcal{L}_{\mathrm{pen}}$ is one of the penalties above and $\lambda_{\text{pen}}$ is a hyperparameter tuning the overall penalty strength. The design is intentionally concise: DPOP preserves the original pairwise loss and adds pressure only on a generated reference response, with a detached gate controlling when that pressure is applied.

\begin{algorithm}[t]
\caption{DPOP loss for one minibatch}
\label{alg:dpop}
\begin{algorithmic}[1]
\STATE Compute $\mathcal{L}_{\mathrm{base}}$ on $(x,y_w,y_l)$.
\STATE Compute $r=\mathrm{sg}(\log \pi_\theta(y_w|x)-\log \pi_\theta(y_l|x))$.
\STATE Set $w_{\mathrm{pen}}=\mathrm{sg}(\mathbf{1}[r<0])f(r)$.
\STATE Evaluate $\mathcal{L}_{\mathrm{pen}}$ on reference-greedy response $y_g$.
\STATE Return $\mathcal{L}_{\mathrm{base}}+\lambda_{\mathrm{pen}} w_{\mathrm{pen}}\mathcal{L}_{\mathrm{pen}}$.
\end{algorithmic}
\end{algorithm}

\section{Experiments}

\subsection{Experimental Setup}
\textbf{Models and data}
We optimize preferences using two widely used instruction-tuned models as the SFT initialization:
\href{https://huggingface.co/meta-llama/Meta-Llama-3-8B-Instruct}
{\texttt{meta-llama/Meta-Llama-3-8B-Instruct}}, and
\href{https://huggingface.co/google/gemma-2-9b-it}
{\texttt{google/gemma-2-9b-it}}.

Training uses UltraFeedback-style preference data \citep{cui2024ultrafeedback}, specifically \url{https://huggingface.co/datasets/princeton-nlp/llama3-ultrafeedback} for the Llama model and \url{https://huggingface.co/datasets/princeton-nlp/gemma2-ultrafeedback-armorm} for the Gemma model.

\textbf{Evaluation benchmarks.}
We evaluate with AlpacaEval 2.0 and report raw win rate (WR), length-controlled win rate (LC-WR), and average response length. We primarily focus on LC-WR because automatic evaluators are known to favor longer responses and length control improves robustness to this confounder \citep{dubois2024length}.

\textbf{Baselines.}
We select SimPO and AlphaDPO as representative baselines. 
While a large body of work has proposed variants of DPO, we found SimPO to be one of the strongest practical baselines in our experimental setting. 
In contrast, several alternative DPO variants did not yield consistent gains over SimPO and were often more sensitive to hyperparameter tuning. 
Therefore, we use SimPO as the primary reference-free baseline and AlphaDPO as a representative margin-based baseline.

\textbf{Computational Environment.} All training experiments were conducted using 8×H100 GPUs, as per the procedures detailed in the alignment-handbook repository.

\begin{figure}[t]
\centering
\includegraphics[width=\linewidth]{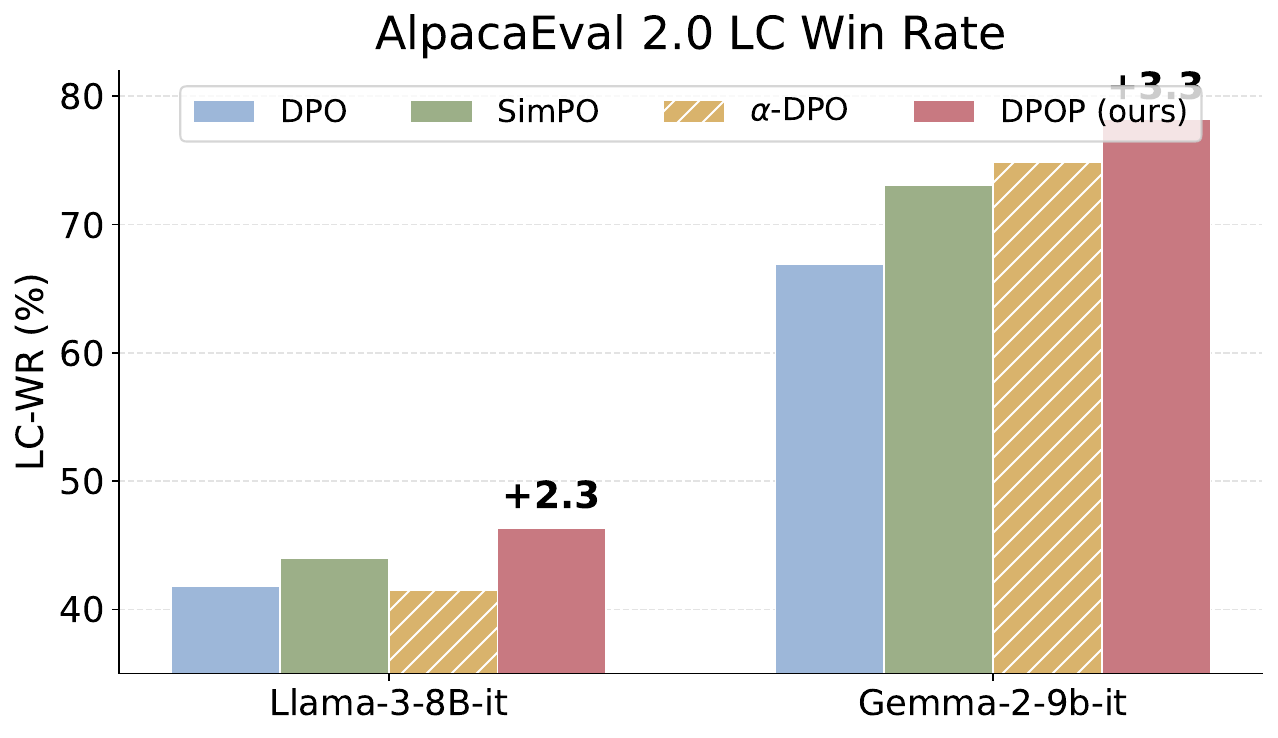}
\caption{AlpacaEval 2.0 length-controlled win rate. DPOP improves over DPO, SimPO, and AlphaDPO on both Llama-3-8b-it and Gemma-2-9b-it. The numbers above DPOP bars are improvements over the best non-DPOP baseline for the same model.}
\label{fig:main_lcwr}
\end{figure}

\begin{table}[t]
\centering
\small
\setlength{\tabcolsep}{4.2pt}
\caption{Main AlpacaEval 2.0 evaluation results. LC-WR is the length-controlled win rate and is the primary metric. The best results are highlighted in bold.}
\label{tab:main_results}
\begin{tabular}{llccc}
\toprule
Model & Method & WR & LC-WR & Length \\
\midrule
Llama-3-8b-it & DPO & 39.49 & 41.84 & 1909 \\
 & SimPO & 37.65 & 44.01 & 1763 \\
 & AlphaDPO & 33.05 & 41.48 & 1675 \\
 & \textbf{DPOP (ours)} & \textbf{47.57} & \textbf{46.35} & 2057 \\
\midrule
Gemma-2-9b-it & DPO & 70.40 & 66.90 & 1900 \\
 & SimPO & 66.12 & 73.08 & 1754 \\
 & AlphaDPO & 65.32 & 74.90 & 1682 \\
 & \textbf{DPOP (ours)} & \textbf{73.76} & \textbf{78.22} & 1970 \\
\bottomrule
\end{tabular}
\end{table}

\subsection{DPOP Consistently Outperforms Preference Optimization Baselines}

\Cref{fig:main_lcwr,tab:main_results} summarize the main comparison. 
Across both Llama-3-8b-it and Gemma-2-9b-it, DPOP achieves the strongest LC-WR among all evaluated methods (DPO, SimPO, and AlphaDPO), suggesting that the reference-greedy response provides useful additional signal beyond the original preference pair. 

We note that the improvement in raw win rate can partly reflect changes in response length, especially on Llama-3-8b-it. 
Therefore, we emphasize LC-WR as the primary metric for comparison. 
Overall, these results support our hypothesis that incorporating the reference-greedy response leads to more effective preference optimization.

\subsection{Comparing Penalty Designs}

\begin{table}[t]
\centering
\small
\setlength{\tabcolsep}{4.2pt}
\caption{Penalty search on Llama-3-8b-it. For each penalty family, we report the best hyperparameters and evaluation results after a grid search.}
\label{tab:penalty_ablation}
\begin{tabular}{lcccccc}
\toprule
Penalty & $\lambda_{\text{pen}}$ & $\beta_p$ & $\gamma_p$ & WR & LC-WR & Length \\
\midrule
SimNPO & 1.0 & 1.0 & 1.0 & 47.57 & \textbf{46.35} & 2057 \\
NPO & 0.01 & 0.05 & -- & 42.76 & 43.94 & 1960 \\
Unlikelihood & 0.01 & -- & -- & 40.07 & 41.36 & 1948 \\
\bottomrule
\end{tabular}
\end{table}

\Cref{tab:penalty_ablation} compares penalty families on Llama-3-8b-it. SimNPO is the strongest variant, reaching 46.35 LC-WR. NPO improves over the DPO baseline but remains below SimNPO at 43.94 LC-WR, while unlikelihood reaches 41.36 LC-WR, performing worse than DPO without penalization.

\subsection{SimNPO Hyperparameter Search}

\begin{table}[t]
\centering
\small
\setlength{\tabcolsep}{3.6pt}
\caption{Top SimNPO-penalty DPOP hyperparameter settings by AlpacaEval 2.0 LC-WR. The sweep varies the penalty weight $\lambda$, penalty temperature $\beta_p$, and SimNPO margin $\gamma_p$.}
\label{tab:simnpo_grid}
\begin{tabular}{llccccc}
\toprule
Model & $\lambda$ & $\beta_p$ & $\gamma_p$ & WR & LC-WR & Length \\
\midrule
Llama-3-8b-it & 1.0 & 1.0 & 1.0 & 47.57 & \textbf{46.35} & 2057 \\
 & 1.0 & 1.0 & 0.0 & 44.77 & 45.14 & 1999 \\
 & 2.0 & 1.0 & 0.0 & 47.08 & 44.47 & 2121 \\
 & 1.0 & 2.0 & 1.0 & 44.78 & 44.28 & 2016 \\
 & 1.0 & 2.0 & 2.0 & 45.21 & 43.52 & 2074 \\
\midrule
Gemma-2-9b-it & 2.0 & 1.0 & 0.0 & 73.76 & \textbf{78.22} & 1970 \\
 & 4.0 & 1.0 & 0.0 & 74.13 & 77.47 & 1999 \\
 & 2.0 & 1.0 & 1.0 & 74.01 & 77.22 & 2014 \\
 & 1.0 & 1.0 & 1.0 & 72.88 & 77.22 & 1981 \\
 & 4.0 & 1.0 & 1.0 & 74.69 & 76.84 & 2073 \\
\bottomrule
\end{tabular}
\end{table}

\Cref{tab:simnpo_grid} summarizes the top SimNPO-penalty settings for both model families. On Llama-3-8b-it, the best setting uses $\lambda=1.0$, $\beta_p=1.0$, and $\gamma_p=1.0$. On Gemma-2-9b-it, the best setting uses $\lambda=2.0$, $\beta_p=1.0$, and $\gamma_p=0.0$. In both cases, the top several configurations remain above their corresponding SimPO and AlphaDPO baselines, indicating that the observed gains are not isolated to a single brittle setting.

\subsection{Weight Function Search}

\begin{table}[t]
\centering
\small
\setlength{\tabcolsep}{4.2pt}
\caption{Weight-function search for $f(r)$ on Llama-3-8b-it.}
\label{tab:f_weight_search}
\begin{tabular}{lccccc}
\toprule
$f(r)$ & WR & LC-WR & Length \\
\midrule
linear  & 47.57 & \textbf{46.35} & 2057 \\
constant   & 47.27 & 45.16 & 1958 \\
square root   & 45.95 & 44.17 & 1984 \\
quadratic  & 45.27 & 45.12 & 1925 \\
\bottomrule
\end{tabular}
\end{table}

\Cref{tab:f_weight_search} compares the four choices for the detached weight function $f(r)$. Among these runs, linear weighting gives the strongest LC-WR, followed closely by the best constant-weight setting. 

The best setting makes the penalty strength proportional to the severity of the policy-margin error. When the policy is only mildly wrong, the rejected response has a small advantage over the chosen response and the additional penalty on the reference-greedy response remains modest. When the policy is strongly biased toward the rejected response, the margin violation is larger and the penalty weight increases accordingly. Thus, linear weighting concentrates optimization pressure on examples where the current policy most clearly fails to represent the observed preference, while avoiding the uniform pressure of a constant weight. The weaker squareroot and quadratic variants suggest that overly compressing or over-amplifying the margin signal is less effective than preserving this direct proportional relationship.

\section{Conclusion}

DPOP extends offline direct preference optimization by using the reference model's own greedy response as a selectively penalized third response. 
The method keeps the original pairwise DPO loss, adds a gated penalty only when the policy margin is still wrong, and uses detached weights to avoid turning the gate into a learned reward signal. 
On AlpacaEval 2.0, DPOP improves LC-WR over DPO, SimPO, and AlphaDPO on both Llama-3-8B-Instruct and Gemma-2-9B-IT, with the strongest results coming from the SimNPO-style penalty.

Future work should evaluate DPOP on a broader set of instruction-following, reasoning, factuality, and safety benchmarks to verify that penalizing reference-greedy responses does not degrade other model capabilities. 
It would also be important to better characterize when reference-greedy responses correspond to harmful local modes that should be penalized, versus acceptable responses that should be preserved.
\bibliography{main}
\bibliographystyle{icml2026}

\end{document}